\documentclass{article}

\usepackage{microtype}
\usepackage{graphicx}
\usepackage{subcaption}
\usepackage{booktabs} 
\usepackage{hyperref}

\usepackage[preprint]{icml2026}

\usepackage{amsmath}
\usepackage{amssymb}
\usepackage{mathtools}
\usepackage{amsthm}

\usepackage[capitalize,noabbrev]{cleveref}

\theoremstyle{plain}

\theoremstyle{definition}

\theoremstyle{remark}

\usepackage[textsize=tiny]{todonotes}

\usepackage{xspace}
\usepackage{colortbl}
\usepackage{multirow}
\usepackage{xcolor}
\usepackage[most]{tcolorbox}
\usepackage{tabularx}
\usepackage{listings}
\usepackage{enumitem}

\usepackage{times}
\usepackage{latexsym}
\usepackage{natbib}

\renewcommand{\algorithmiccomment}[1]{\hfill $\triangleright$ #1}
\newcommand{\STEP}[1]{\STATE \textit{\# #1}}

\newcommand{\method}{Low-Rank Decomposed Scaling\xspace}
\newcommand{\methodFull}{Low-Rank Decomposed Scaling (LoRDS)\xspace}
\newcommand{\methodShort}{LoRDS\xspace}

\icmltitlerunning{Breaking the Blocks: Continuous Low-Rank Decomposed Scaling for Unified LLM Quantization and Adaptation}

\begin{document}

\twocolumn[
  \icmltitle{Breaking the Blocks: Continuous Low-Rank Decomposed Scaling for \\ Unified LLM Quantization and Adaptation}
  
  \icmlsetsymbol{equal}{*}

  \begin{icmlauthorlist}
    \icmlauthor{Pingzhi Tang}{equal,iai,yuanpei}
    \icmlauthor{Ruijie Zhou}{equal,hit}
    \icmlauthor{Fanxu Meng}{iai,bigai}
    \icmlauthor{Wenjie Pei}{hit}
    \icmlauthor{Muhan Zhang}{iai,bigai}
  \end{icmlauthorlist}

  \icmlaffiliation{iai}{Institute for Artificial Intelligence, Peking University, Beijing, China}
  \icmlaffiliation{yuanpei}{Yuanpei College, Peking University, Beijing, China}
  \icmlaffiliation{bigai}{State Key Laboratory of General Artificial Intelligence, BIGAI, Beijing, China}
  \icmlaffiliation{hit}{Harbin Institute of Technology, Shenzhen, China}

  \icmlcorrespondingauthor{Muhan Zhang}{muhan@pku.edu.cn}

  \icmlkeywords{Machine Learning, ICML}

  \vskip 0.3in
]

\printAffiliationsAndNotice{\icmlEqualContribution}

\begin{abstract}
Current quantization methods for LLMs predominantly rely on block-wise structures to maintain efficiency, often at the cost of representational flexibility.
In this work, we demonstrate that element-wise quantization can be made as efficient as block-wise scaling while providing strictly superior expressive power by modeling the scaling manifold as continuous low-rank matrices ($S = BA$). We propose \methodFull, a unified framework that rethinks quantization granularity through this low-rank decomposition.
By ``breaking the blocks'' of spatial constraints, \methodShort establishes a seamless efficiency lifecycle: it provides high-fidelity PTQ initialization refined via iterative optimization, enables joint QAT of weights and scaling factors, and facilitates high-rank multiplicative PEFT adaptation. Unlike additive PEFT approaches such as QLoRA, \methodShort enables high-rank weight updates within a low-rank budget while incurring no additional inference overhead.
Supported by highly optimized \texttt{Triton} kernels, \methodShort consistently outperforms state-of-the-art baselines across various model families in both quantization and downstream fine-tuning tasks.
Notably, on Llama3-8B, our method achieves up to a 27.0\% accuracy improvement at 3 bits over NormalFloat quantization and delivers a 1.5$\times$ inference speedup on NVIDIA RTX 4090 while enhancing PEFT performance by 9.6\% on downstream tasks over 4bit QLoRA, offering a robust and integrated solution for unified compression and adaptation of LLMs.
\end{abstract}
\newpage
\section{Introduction}
\label{sec:intro}

\begin{figure*}[t]
    \centering
    \begin{subfigure}[b]{0.58\textwidth}
        \centering
        \includegraphics[width=0.9\textwidth]{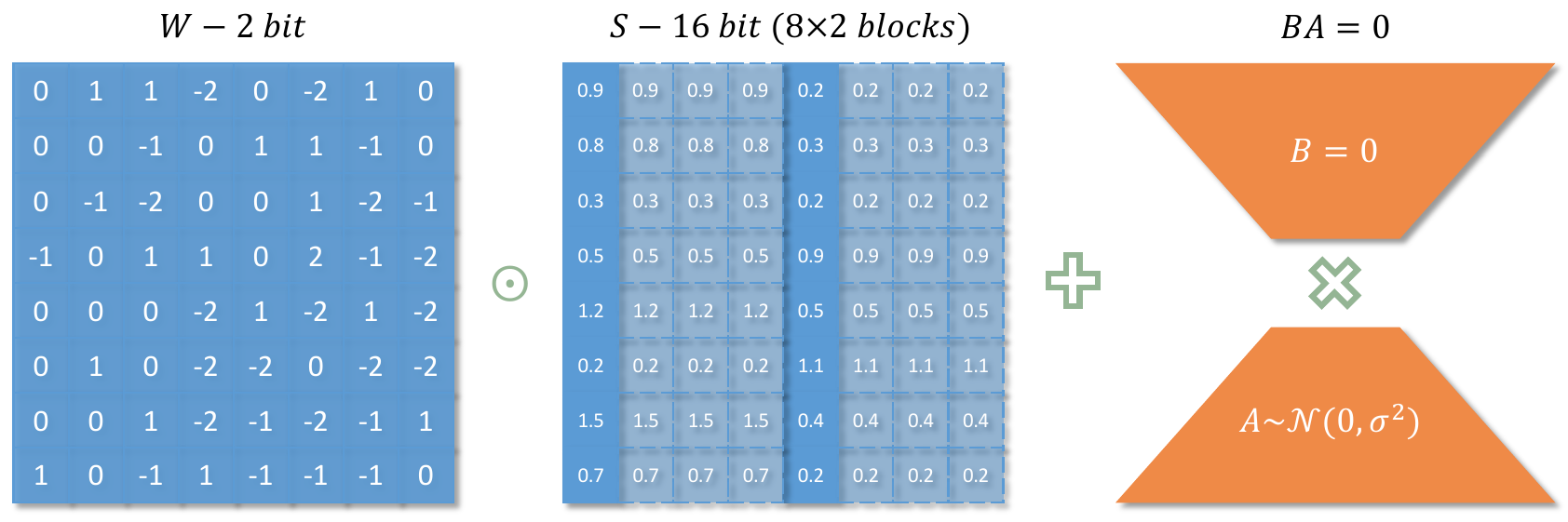}
        \caption{QLoRA}
        \label{fig:qlora}
    \end{subfigure}
    \hfill
    \begin{subfigure}[b]{0.38\textwidth}
        \centering
        \includegraphics[width=0.9\textwidth]{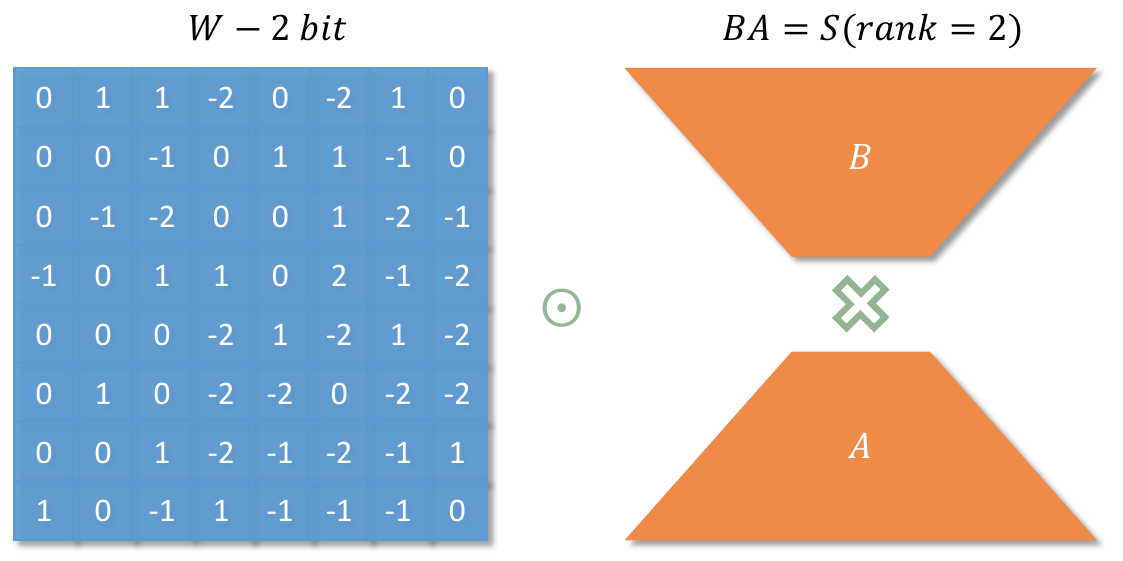}
        \caption{\methodShort}
        \label{fig:lords}
    \end{subfigure}
    \caption{Architectural comparison between (a) QLoRA and (b) our proposed \methodShort. (a) QLoRA employs standard block-wise quantization where parameters within each block share a static scaling factor. During fine-tuning, it introduces an auxiliary, full-precision additive adapter ($W + BA$), which cannot be merged into the quantized weights, leading to additional inference latency.
    (b) LoRDS fundamentally rethinks the scaling mechanism by decomposing the block-wise scaling matrix $S$ into a low-rank product $BA$. This multiplicative formulation ($W \odot BA$) serves a dual purpose: it can be refined via PTQ or QAT to achieve more granular scaling, or directly fine-tuned for PEFT. \methodShort enables high-rank parameter updates and incurs zero additional inference overhead, as the learned scaling factors are naturally absorbed into the dequantization process.}
    \label{fig:hero}
\end{figure*}

Large Language Models (LLMs, \citealt{brown2020language, vaswani2017attention}) have demonstrated remarkable capabilities across a wide range of natural language processing tasks, fundamentally reshaping the landscape of artificial intelligence. However, their exceptional capabilities come at significant resource demands: the immense parameter counts impose large memory footprints and computational costs, making both efficient \textit{deployment} and downstream \textit{adaptation} significant challenges. 
To mitigate these bottlenecks, extensive research has been dedicated to model compression and efficient tuning. 
\textit{Model Quantization}~\citep{shen2020q, zafrir2019q8bert, bai2022towards, dettmers2022gpt3, banner2019post, choukroun2019low} reduces memory usage by approximating high-precision weights with low-bit numbers, significantly lowering storage requirements. 
\textit{Parameter-Efficient Fine-Tuning} (PEFT, \citealt{xu2023parameter, han2024parameter}) methods, such as LoRA~\citep{hu2022lora}, freeze the pretrained model and update only a small set of auxiliary parameters, thereby reducing training costs. 
Efforts have also been made to integrate these two paradigms.
Methods like QLoRA~\citep{dettmers2023qlora} and LoftQ~\citep{li2023loftq} combine quantization with LoRA adapters, PEQA~\citep{kim2023memory} directly finetunes the quantization parameters.

Despite these advancements, current approaches face several critical limitations. 
First, the dominant quantization paradigm relies heavily on \textit{block-wise quantization}~\citep{dettmers20218}, where weights are grouped into blocks that share quantization parameters. While this approach implicitly assumes that neighboring parameters possess similar statistical properties, this assumption lacks theoretical guarantees and often fails to capture weight distributions, especially in the presence of outliers.
Second, the landscape of efficient LLM utilization remains fragmented: Post-Training Quantization (PTQ, \citealt{frantar2022gptq,lin2024awq}), Quantization-Aware Training (QAT, \citealt{gupta2015deep, jacob2018quantization, krishnamoorthi2018quantizing, liu2024llm, esser2019learned, bhalgat2020lsq+}), and Quantized PEFT are typically treated as distinct optimization problems with separate methodologies, lacking a unified framework.
Finally, existing quantized PEFT methods like QLoRA introduce additive adapters alongside quantized weights. Unlike full-precision fine-tuning where adapters can be merged, the mixed-precision nature of QLoRA prevents lossless merging, incurring additional latency and computational overhead during inference.

To address these challenges, we propose \methodFull, a unified framework that rethinks the granularity of quantization scaling. 
\methodShort improves upon traditional block-wise quantization by breaking the rigid block structure. Instead of independent scalars for each block, we model the scaling factors as a continuous low-rank matrix decomposed via Singular Value Decomposition (SVD). 
This approach effectively occupies a representational space between coarse \textit{per-block scaling} and dense \textit{per-element scaling}; while \methodShort can be seamlessly initialized from standard block-wise statistics, its continuous low-rank formulation offers strictly superior expressive power.
This formulation allows for global coherence and more flexible representations of the weight magnitude distribution.
Crucially, \methodShort unifies the lifecycle of LLM efficiency:
(1) As a \textbf{PTQ} method, it can be initialized from standard block-wise settings and refined through low-cost alternating optimization, achieving significantly lower quantization error.
(2) As a \textbf{QAT} method, the full-precision weights and the decomposed scaling matrices ($A$ and $B$) can be trained jointly using Straight-Through Estimators (STE), pushing the limits of performance.
(3) As a \textbf{PEFT} method, by fine-tuning only the low-rank scaling matrices $A$ and $B$, \methodShort achieves superior performance compared to additive methods like QLoRA and LoftQ. 
Crucially, unlike LoRA-based methods that are restricted to low-rank additive updates, the multiplicative nature of \methodShort enables effectively high-rank parameter updates.
Furthermore, as the adaptation is intrinsic to the dequantization scaling process, \methodShort does not incur additional inference overhead compared to standard quantized models.

We extensively evaluate \methodShort across various model families (e.g., LLaMA-3, Qwen-3) and downstream tasks. 
Experimental results demonstrate that \methodShort consistently minimizes quantization errors in both PTQ and QAT settings by leveraging its continuous low-rank decomposed scaling mechanism. Furthermore, thanks to its ability to facilitate high-rank weight updates, \methodShort achieves superior adaptation performance in PEFT tasks while maintaining a smaller floating-point parameter budget than existing baselines. Crucially, our approach eliminates the inference bottleneck inherent in additive adapters, delivering up to a 1.5$\times$ inference speedup compared to QLoRA on NVIDIA RTX 4090 GPUs.

In summary, our main contributions are as follows:
\begin{itemize}[leftmargin=*, noitemsep, topsep=0pt]
    \item We propose \methodFull, a novel quantization framework that interpolates between coarse per-block and dense per-element scaling with learnable low-rank matrices, effectively capturing complex weight distributions.
    \item We present a unified formulation that seamlessly covers PTQ, QAT, and PEFT, offering a versatile solution for the entire LLM efficiency lifecycle.
    \item We achieve state-of-the-art performance in both quantization reconstruction and downstream fine-tuning tasks, outperforming existing additive PEFT methods without introducing any inference latency.
    \item We implement highly optimized \texttt{Triton} kernels for inference of \methodShort, bridging the gap between theoretical efficiency and real-world deployment speed.
\end{itemize}
\section{Related Work}
\label{sec:related_work}
\textbf{Post-Training Quantization (PTQ)} is widely adopted due to its ability to be directly applied to pretrained LLMs, typically with the aid of a small calibration dataset.
PTQ methods are commonly divided into weight-only quantization and weight-activation quantization.
Weight-only quantization focuses on compressing model parameters into low-bit representations.
GPTQ~\citep{frantar2022gptq} leverages second-order information for accurate layer-wise quantization.
AWQ \citep{lin2024awq} identifies salient weight channels based on activation distributions and protects them through equivalent transformations.
Quip\#~\citep{quip-sharp} employs Hadamard-based incoherence and structured codebooks to improve robustness under ultra-low-bit quantization.
Weight-activation quantization~\citep{xiao2023smoothquant,odyssey,quarot,dong2025ditas} further compresses both weights and activations.
SmoothQuant mitigates activation outliers by smoothing scale distributions.
Odyssey presents a deployment-oriented PTQ framework that improves inference efficiency.
DITAS enhances activation smoothing to enable effective quantization of diffusion transformers.
Our work shares a conceptual direction with LRQ~\citep{lee2025lrq} in utilizing low-rank approximations for quantization scaling factors. However, while LRQ employs an exponential mapping that precludes direct initialization and necessitates a computationally intensive reconstruction process from scratch, our linear formulation ($S = BA$) enables a seamless transition from standard block-wise statistics via SVD with minimal refinement. Furthermore, while LRQ is specifically designed for PTQ, \methodShort provides a unified framework encompassing PTQ, QAT, and quantized PEFT, facilitating downstream adaptation through a trainable scaling representation that remains inaccessible in the LRQ formulation.

\textbf{Quantization-Aware Training (QAT)} explicitly models quantization effects during training, enabling higher accuracy than PTQ under aggressive low-bit settings.
LLM-QAT~\citep{liu2024llm} pioneered the exploration of QAT specifically for large language models, demonstrating the feasibility of fine-tuning quantized weights to recover performance. 
Building on this, BitDistiller~\citep{bitdistiller} incorporates knowledge distillation into the QAT process to further improve training stability and performance.
OneBit~\citep{xu2024onebit} employs QAT to achieve extreme binary and ternary quantization.
SpinQuant~\citep{liu2024spinquant} introduces learned rotation matrices to optimize quantized networks and improve robustness under low-bit precision.

\textbf{Parameter-efficient fine-tuning (PEFT)} aims to reduce the computational cost of adapting large language models by updating only a small subset of parameters.
LoRA~\citep{hu2022lora} popularized this paradigm by introducing trainable low-rank additive matrices alongside frozen pretrained weights.
QLoRA~\citep{dettmers2023qlora} further quantizes the frozen weights that are only required for forward propagation, substantially lowering memory and computation overhead during training.
To mitigate the quantization-induced errors introduced by quantized fine-tuning, LoftQ~\citep{li2023loftq} and QPiSSA~\citep{meng2024pissa} propose strategies to restore more information in adapters, providing better initializations.
FALQON~\citep{choi2025falqon} and QA-LoRA~\citep{xu2023qa} improve computational efficiency in both the fine-tuning and inference stages by jointly considering quantization and low-rank adaptation.
Beyond simple additive updates, HiRA \citep{huang2025hira} pioneers the use of parameter-efficient Hadamard high-rank adaptation, discovering that capturing high-rank weight updates is instrumental for effective model adaptation.

\section{\method}
\label{sec:method}

In this section, we present \methodFull, a unified framework that reimagines quantization scaling through the lens of continuous low-rank approximation. 
In the following, we first formalize conventional per-block quantization and then introduce our low-rank scaling mechanism and its integrations with Post-Training Quantization (PTQ), Quantization-Aware Training (QAT), and Quantized Parameter-Efficient Fine-Tuning (PEFT).

\subsection{Preliminaries}
\label{method:preliminary}

\textbf{Block-wise Quantization}
(or group-wise quantization) partitions a weight matrix $W \in \mathbb{R}^{n \times m}$ into $K = (nm)/B$ contiguous blocks of size $B$, normalizing each independently. In this work, we focus on symmetric quantization where each block $W_b$ is characterized by a scaling factor $s_b$. 
The quantized representation $Q_b$ and the corresponding de-quantized weight $\widehat{W}_b$ are then obtained via:
\begin{gather} 
Q_b = \text{\textsc{Quant}}(W_b, s_b) = \text{\textsc{Round}} \left( \frac{W_b}{s_b} \right), \\
\widehat{W}_b = \text{\textsc{Dequant}}(Q_b, s_b) = Q_b \cdot s_b, 
\end{gather}
where $\text{\textsc{Round}}(\cdot)$ denotes the mapping to the nearest discrete levels of a target data type, such as INT4 or NF4~\citep{dettmers2023qlora}.

For a matrix-centric formulation, we define a global scaling matrix $S \in \mathbb{R}^{n \times m}$ constructed by repeating the block scales $s \in \mathbb{R}^{n \times (m/B)}$ across each block, such that $S = s \otimes \mathbf{1}_{1 \times B}$. The full-matrix quantization is then expressed as $\widehat{W} = \text{\textsc{Round}}(W \oslash S) \odot S$, where $\oslash$ and $\odot$ denote Hadamard division and multiplication. Importantly, this block-wise scaling matrix is inherently redundant with $rank(S) \leq m/B$. Furthermore, this paradigm serves as a unified framework for various granularities: per-row ($B=m$), per-column ($B=n$), and per-tensor scaling are all special cases of the block-wise structure.

\paragraph{QAT and Straight-Through Estimator (STE)} 
While Post-Training Quantization (PTQ) focuses on minimizing reconstruction error without retraining, QAT integrates quantization into the training loop to allow the model to adapt to quantization noise. However, the rounding operation is non-differentiable, yielding zero gradients almost everywhere. To overcome this, the Straight-Through Estimator (STE, \citealt{bengio2013estimating, liu2024llm}) is typically employed, which approximates the gradient of the rounding function as an identity mapping during the backward pass, i.e., $\partial \text{\textsc{Round}}(x) / \partial x \approx 1$.

\subsection{Low-Rank Scaling Decomposition}
\label{method:low-rank}

While block-wise quantization significantly reduces the overhead of scaling factors, its rigid structure forces all parameters within a block to share a single scalar. This heuristic often fails to capture the true heterogeneity of weight distributions, especially in the presence of outliers. Ideally, one would desire per-element scaling to achieve maximum precision, yet the parameter cost is prohibitive. \textbf{\methodShort bridges this gap by positioning its representational capacity between coarse per-block and dense per-element scaling}: it achieves the expressive flexibility of the latter while maintaining the parameter efficiency of the former through low-rank decomposition.

As established in Section \ref{method:preliminary}, the block-wise scaling matrix $S \in \mathbb{R}^{n \times m}$ is inherently redundant and low-rank. We propose to decompose $S$ into a continuous low-rank manifold via truncated Singular Value Decomposition (SVD):
\begin{equation}
S \approx U_r \Sigma_r V_r^T = (U_r \Sigma_r^{1/2})(\Sigma_r^{1/2} V_r^T) = BA,
\end{equation}
where $B \in \mathbb{R}^{n \times r}$ and $A \in \mathbb{R}^{r \times m}$. While this initialization exactly recovers the original block-wise statistics, the factorization $BA$ offers strictly superior expressiveness as it is no longer restricted to a piecewise-constant structure. To ensure a rigorous comparison, we align our scaling parameter budget $r(n+m)$ with the block-wise budget $nm/B$ by setting the equivalent rank $r = \lfloor \frac{nm}{B(n+m)} \rfloor$ (see Appendix~\ref{app:rank} for details). Consequently, \methodShort captures fine-grained, near per-element weight characteristics within a standard per-block parameter budget.

\subsection{Refining the Scaling Manifold via PTQ and QAT}
\label{method:quantization}

The low-rank scaling decomposition provides a flexible initialization that preserves block-wise accuracy while facilitating a higher-dimensional space for fine-grained adjustments. We leverage this potential through both PTQ and QAT.

\paragraph{Iterative PTQ Refinement}
As a PTQ method, \methodShort refines the initial SVD decomposition to further minimize the Frobenius reconstruction error: $\min_{B, A, Q} \| W - (BA) \odot Q \|_F^2$. We adopt an alternating optimization strategy, as detailed in Algorithm~\ref{alg:refinement}.

\begin{algorithm}[tb]
  \caption{\methodShort PTQ Optimization}
  \label{alg:refinement}
  \small
  \begin{algorithmic}
    \REQUIRE Weight $W \in \mathbb{R}^{n \times m}$, target rank $r$, iteration steps $T$, learning rate $\eta$, target precision look-up table $\mathcal{L}$.
    \ENSURE Optimized low-rank matrices $B \in \mathbb{R}^{n \times r}, A \in \mathbb{R}^{r \times m}$, and quantized matrix $Q \in \mathbb{R}^{n \times m}$.
    
    \STEP{Step 1: Initialization via Truncated SVD}
    \STATE $S \leftarrow \textsc{ComputeScale}(W, \text{block size}=\frac{m}{r})$
    \STATE $U, \Sigma, V^T \leftarrow \text{SVD}(S)$
    \STATE $B \leftarrow U_{:, 1:r} \Sigma_{1:r, 1:r}^{1/2}$ \COMMENT{Initialize $B, A$ using SVD}
    \STATE $A \leftarrow \Sigma_{1:r, 1:r}^{1/2} V_{1:r, :}^T$
    
    \STEP{Step 2: Iterative Refinement}
    \FOR{$t=1$ {\bfseries to} $T$}
      \STEP{Step 2.1: Quantization Step (Fixed $B, A$)}
      \STATE $S \leftarrow B A$ \COMMENT{Compute current scale}
      \FOR{$i=1$ {\bfseries to} $n$}
        \FOR{$j=1$ {\bfseries to} $m$}
          \STATE $Q_{ij} \leftarrow \arg\min_{v \in \mathcal{L}} (S_{ij} \cdot v - W_{ij})^2$
        \ENDFOR
      \ENDFOR
      
      \STEP{Step 2.2: Adaptation Step (Fixed $Q$)}
      \STATE Compute objective: $\mathcal{L}_{\text{MSE}} = \| W - (B A) \odot Q \|_F^2$
      \STATE $B \leftarrow B - \eta \cdot \mathcal{G}_B$ \COMMENT{Update $B, A$ using AdamW}
      \STATE $A \leftarrow A - \eta \cdot \mathcal{G}_A$
    \ENDFOR
    
    \STATE {\bfseries return} $B, A, Q$
  \end{algorithmic}
\end{algorithm}

\begin{table*}[t]
    \caption{Post-Training Quantization (PTQ)  results comparing the proposed LoRDS method against state-of-the-art PTQ techniques, including standard NF4, GPTQ (INT4), AWQ (INT4), and LoftQ (\textbf{which incorporates additional low-rank adapters, rank=16}). \textbf{Wiki} and \textbf{PTB} refer to perplexity ($\downarrow$) on WikiText-2 and Penn Treebank. The remaining columns report zero-shot accuracy ($\uparrow$) on \textbf{BoolQ}, \textbf{PIQA}, \textbf{HS} (HellaSwag), \textbf{WG} (WinoGrande), \textbf{ARC-e} (ARC-Easy), \textbf{ARC-c} (ARC-Challenge), and \textbf{OBQA} (OpenBookQA). }
    \vspace{-5pt}
    \label{tab:ptq_results}
    \begin{center}
    \begin{small}
    \setlength{\tabcolsep}{3.5pt}
    \begin{tabular}{lccccccccccc}
        \toprule
        \textbf{Model} & \textbf{Method} & \textbf{Wiki} $\downarrow$ & \textbf{PTB} $\downarrow$ & \textbf{BoolQ} & \textbf{PIQA} & \textbf{HS} & \textbf{WG} & \textbf{ARC-e} & \textbf{ARC-c} & \textbf{OBQA} & \textbf{Avg} $\uparrow$ \\
        \midrule
        \rowcolor{gray!10} Llama3-8B & - & 7.26 & 15.84 & 81.28 & 79.49 & 60.23 & 73.09 & 80.13 & 50.43 & 34.80 & 65.64 \\
        \midrule
        \multirow{5}{1.5cm}{\centering block size 128}
         & NF4 & 7.90 & 16.85 & 80.21 & 78.13 & 59.12 & 73.09 & \textbf{80.22} & 48.98 & 34.20 & 64.85 \\
         & GPTQ & 10.01 & 17.21 & 80.55 & \underline{79.22} & 59.13 & \textbf{74.43} & 79.59 & \textbf{49.83} & 33.20 & 65.13 \\
         & AWQ & 8.07 & 17.10 & 75.29 & \textbf{79.49} & \underline{59.21} & 72.22 & 78.75 & \underline{49.40} & 34.00 & 64.05 \\
         & LoftQ & \underline{7.86} & \underline{16.75} & \underline{81.01} & 78.84 & 59.01 & \underline{73.88} & 79.38 & 48.72 & \textbf{35.40} & \underline{65.18} \\
         & \methodShort & \textbf{7.77} & \textbf{16.54} & \textbf{82.08} & 79.11 & \textbf{59.30} & 73.40 & \underline{79.88} & \underline{49.40} & \underline{34.40} & \textbf{65.37} \\
        \cmidrule{2-12}
        \multirow{5}{1.5cm}{\centering block size 256}
         & NF4 & 7.98 & 16.96 & \underline{81.04} & 78.51 & 58.82 & \underline{73.56} & 78.87 & \underline{48.89} & 32.00 & 64.53 \\
         & GPTQ & 8.44 & 17.17 & 78.17 & \underline{78.78} & 57.09 & 73.24 & 78.66 & 46.59 & \textbf{33.60} & 63.73 \\
         & AWQ  & 8.30 & 17.40 & 74.19 & 77.97 & 58.67 & 72.22 & 78.03 & 47.70 & \textbf{33.60} & 63.20 \\
         & LoftQ & \underline{7.94} & \underline{16.88} & \textbf{81.35} & 78.02 & \underline{58.97} & 73.09 & \underline{79.42} & 48.81 & 33.00 & \underline{64.66} \\
         & \methodShort & \textbf{7.81} & \textbf{16.63} & 80.55 & \textbf{79.22} & \textbf{59.13} & \textbf{74.43} & \textbf{79.59} & \textbf{49.83} & \underline{33.20} & \textbf{65.13} \\
        \midrule
        \rowcolor{gray!10} Qwen3-8B & - & 12.22 & 25.06 & 86.61 & 76.33 & 57.14 & 68.43 & 83.46 & 55.55 & 31.20 & 65.53 \\
        \midrule
        \multirow{5}{1.5cm}{\centering block size 128}
         & NF4 & 12.73 & 26.66 & \underline{86.12} & \underline{76.28} & 55.87 & 67.32 & 82.49 & 53.07 & \textbf{33.20} & 64.91 \\
         & GPTQ & \underline{12.63} & \underline{26.45} & 86.09 & \textbf{76.44} & \underline{56.41} & 65.11 & \underline{82.83} & 54.69 & 29.20 & 64.40 \\
         & AWQ & 12.84 & 27.15 & 85.50 & 75.68 & 55.66 & 67.48 & 81.31 & 54.95 & 31.20 & 64.54 \\
         & LoftQ & 12.66 & 26.49 & 85.78 & 76.06 & 56.04 & \textbf{68.27} & 82.53 & \underline{55.72} & \underline{32.20} & \underline{65.23} \\
         & \methodShort & \textbf{12.29} & \textbf{25.48} & \textbf{86.48} & \underline{76.28} & \textbf{56.58} & \underline{67.96} & \textbf{82.87} & \textbf{56.48} & 31.80 & \textbf{65.49} \\
        \cmidrule{2-12}
        \multirow{5}{1.5cm}{\centering block size 256}
         & NF4 & 12.64 & \underline{25.90} & \underline{86.12} & 76.28 & \underline{55.72} & 67.88 & \underline{82.07} & 52.99 & \underline{32.00} & \underline{64.72} \\
         & GPTQ & 12.63 & 26.10 & 85.93 & \textbf{76.55} & \textbf{56.19} & \textbf{68.27} & 80.72 & 51.28 & 31.20 & 64.31 \\
         & AWQ & 13.48 & 28.80 & 85.81 & \underline{76.39} & \underline{55.72} & 65.98 & 80.77 & 52.30 & \textbf{33.40} & 64.34 \\
         & LoftQ & \underline{12.57} & \underline{25.90} & 85.90 & 76.01 & 55.71 & 67.09 & 82.03 & \underline{53.67} & 31.80 & 64.60 \\
         & \methodShort & \textbf{12.53} & \textbf{25.81} & \textbf{86.30} & 75.63 & \textbf{56.19} & \underline{68.11} & \textbf{83.00} & \textbf{55.55} & 29.80 & \textbf{64.94} \\
        \midrule
        \rowcolor{gray!10} Qwen3-4B & - & 16.45 & 37.77 & 85.02 & 74.86 & 52.26 & 65.90 & 80.30 & 50.60 & 29.80 & 62.68 \\
        \midrule
        \multirow{5}{1.5cm}{\centering block size 128}
         & NF4 & 17.50 & 42.60 & 84.25 & 73.29 & 51.02 & \underline{64.01} & 77.95 & 46.84 & \underline{29.60} & 60.99 \\
         & GPTQ & \underline{17.37} & 41.11 & 84.31 & 73.72 & \textbf{51.19} & 61.48 & \underline{78.87} & \textbf{49.66} & \textbf{30.20} & 61.35 \\
         & AWQ & 17.48 & \underline{40.97} & 83.06 & \textbf{75.08} & 50.56 & 63.93 & 77.99 & 46.59 & 28.80 & 60.86 \\
         & LoftQ & \textbf{17.18} & \textbf{40.05} & \textbf{84.65} & 73.78 & 50.98 & 63.22 & \textbf{79.55} & \underline{49.40} &\underline{ 29.60} & \underline{61.60} \\
         & \methodShort & 17.31 & 45.76 & \underline{84.56} & \underline{74.37} & \underline{51.05} & \underline{64.72} & 78.49 & 48.89 & 29.20 & \textbf{61.61} \\
        \cmidrule{2-12}
        \multirow{5}{1.5cm}{\centering block size 256}
         & NF4 & 17.67 & 43.02 & \underline{84.19} & 73.67 & 50.63 & 62.04 & 76.60 & \underline{48.04} & \underline{30.80} & 60.85 \\
         & GPTQ & \textbf{17.38} & \textbf{40.04} & 83.39 & 73.67 & \underline{50.85} & 62.51 & 75.93 & 47.18 & 30.00 & 60.50 \\
         & AWQ & 17.68 & 42.74 & 81.90 & \underline{73.83} & 50.00 & 61.17 & 76.56 & 45.90 & 28.40 & 59.68 \\
         & LoftQ & \underline{17.39} & \underline{40.97} & 84.13 & 73.78 & 50.67 & \underline{62.75} &\textbf{78.28} & \textbf{48.38} & 30.40 & \underline{61.20} \\
         & \methodShort & 17.50 & 48.68 & \textbf{84.68} & \textbf{74.92} & \textbf{51.30} & \textbf{64.33} & \underline{76.85} & 45.39 & \textbf{31.20} & \textbf{61.24} \\
        \bottomrule
    \end{tabular}
    \vspace{-14pt}
    \end{small}
    \end{center}
\end{table*}

The process consists of two primary phases:(1) a Quantization Step, where $B$ and $A$ are fixed and the optimal discrete levels $Q$ are selected via nearest-neighbor search in the look-up table $\mathcal{L}$; and(2) an Adaptation Step, where $Q$ is fixed and the scaling matrices $B$ and $A$ are updated to minimize the residual error.
Crucially, this optimization is highly efficient. In practice, our PTQ refinement requires less than 30 minutes on a single A100 GPU for an 8B model, making it a practical solution for rapid deployment.

\paragraph{Seamless Integration with QAT}
\methodShort can be seamlessly integrated into the QAT by defining the ``fake-quantized'' weight as a differentiable function of weights and scaling factors: $\widehat{W} = \text{\textsc{Round}}(W \oslash (BA)) \odot (BA)$.
Using STE, we approximate the gradients as:
\begin{gather}
\nabla_W \mathcal{L} = \frac{\partial \mathcal{L}}{\partial \widehat{W}} \cdot \frac{\partial \widehat{W}}{\partial W} \approx \frac{\partial \mathcal{L}}{\partial \widehat{W}} \\
\nabla_S \mathcal{L} = \frac{\partial \mathcal{L}}{\partial \widehat{W}} \cdot \frac{\partial \widehat{W}}{\partial S} \approx \frac{\partial \mathcal{L}}{\partial \widehat{W}} \cdot \left( Q - W \oslash S \right),
\end{gather}
where $S = BA$ and $Q = \text{\textsc{Round}} \left( W \oslash (BA) \right)$.

By jointly fine-tuning the continuous scaling factors and the weights, \methodShort allows the model to better adapt to the discrete constraints, effectively recovering the performance degradation incurred during the quantization process.

\subsection{Quantized Multiplicative Parameter-Efficient Fine-Tuning}

Beyond static quantization, the low-rank scaling decomposition naturally extends to Parameter-Efficient Fine-Tuning (PEFT). Unlike established quantized-PEFT methods like QLoRA which introduce auxiliary full-precision additive adapters ($\Delta W = B_{\text{lora}} A_{\text{lora}}$), \methodShort facilitates adaptation by directly fine-tuning the intrinsic scaling matrices $B$ and $A$, yielding a multiplicative update: $\Delta W = Q \odot (B'A' - BA)$.

A fundamental bottleneck in quantized additive PEFT is the precision mismatch that prevents merging adapters back into quantized weights, inevitably incurring additional latency.
In contrast, since the matrices $B$ and $A$ in \methodShort are intrinsic to the dequantization process, task-specific knowledge is naturally absorbed into the scaling factors. As a result, the post-trained model maintains the same architecture and inference speed as the base quantized model, achieving zero additional inference overhead. 
We further implement optimized \texttt{Triton} kernels to ensure these theoretical gains translate into superior real-world throughput (Section~\ref{exp:kernel}).

The multiplicative nature of \methodShort provides superior expressive power. While standard LoRA is strictly rank-constrained, the element-wise interaction between the high-rank pre-trained matrix $W$ and the low-rank scaling factors $(B'A' - BA)$ effectively unfolds the update across the entire weight manifold. This allows the effective rank of the update to significantly exceed the parameter budget $r$, enabling the model to capture fine-grained adaptation patterns that are inaccessible to additive counterparts~\citep{huang2025hira, meng2024clover}.

\section{Experiments}
\label{sec:experiments}

We evaluate \methodShort across the LLM efficiency lifecycle, focusing on four key dimensions: (1) PTQ Reconstruction: benchmark 4-bit and low-bit performance and the efficacy of our iterative refinement (Section~\ref{exp:ptq}); (2) QAT Integration: assess gains over standard INT4 pipelines (Section~\ref{exp:qat}); (3) Quantized-PEFT: investigate adaptation capabilities (Section~\ref{exp:peft}); and (4) Hardware Efficiency: benchmark custom \texttt{Triton} kernels for real-world throughput (Section~\ref{exp:kernel}).

\subsection{\methodShort for Post-Training Quantization}
\label{exp:ptq}

\textbf{Standard 4-bit Quantization Performance  }
We first compare \methodShort against state-of-the-art PTQ methods, including standard NF4~\citep{dettmers2023qlora}, GPTQ~\citep{frantar2022gptq}, AWQ~\citep{lin2024awq}, and LoftQ~\citep{li2023loftq}, across LLaMA3-8B, Qwen3-8B, and Qwen3-4B models with (equivalent) block sizes of 128 and 256. Evaluation metrics include perplexity (PPL) on WikiText-2 (Wiki, \citealp{merity2016pointer}) and Penn Treebank (PTB, \citealp{marcus1993building}), alongside zero-shot accuracy on a diverse suite of commonsense reasoning tasks.
We set \methodShort ranks to ensure strict parameter parity. Notably, LoftQ (rank 16, 5 iterations) uses ~40M more high-precision parameters compared to other baselines. As shown in Table \ref{tab:ptq_results}, \methodShort leads across all configurations; e.g., on Llama3-8B ($B=256$), \methodShort (65.13\%) surpasses NF4 (64.53\%) and outperforms LoftQ (64.66\%) despite the latter requires an additional $\sim$40 million parameters.

\textbf{Effectiveness of Iterative Refinement  }
A key advantage of our continuous low-rank formulation over rigid block-wise scaling is the ability to perform iterative refinement to minimize reconstruction error. To validate this mechanism, we define the quantization error as the nuclear norm of the quantization residual ($\| W - Q \|_*$), averaged over all modules, and monitor it along with perplexity and accuracy before and after refinement over 500 steps using a learning rate of 0.05.
The results in Table~\ref{tab:iter_effectiveness} confirm that our optimization effectively minimizes reconstruction error.
For Llama3-8B ($B=256$), the iterative process reduces the quantization error from 357.95 (initial SVD) to 329.39, leading to a significant drop in WikiText-2 PPL from 8.28 to 7.81 and a 0.6\% accuracy gain. This confirms the effectiveness of our proposed refinement process.

\begin{table}[ht]
\caption{Impact of iterative refinement on quantization error and perplexity on various models and block sizes. \textbf{Iter.} indicates whether the iterative refinement process is applied. \textbf{QuantError} is measured as the nuclear norm of the quantization residual.}
\vspace{-5pt}
\label{tab:iter_effectiveness}
\begin{center}
\begin{small}
\setlength{\tabcolsep}{3.5pt}
\begin{tabular}{lccccc}
\toprule
\textbf{Model} & \textbf{BlockSize} & \textbf{Iter.} & \textbf{QuantError} $\downarrow$ & \textbf{Wiki} $\downarrow$ & \textbf{Avg} $\uparrow$ \\
\midrule
\multirow{4}{*}[-2pt]{Llama3-8B} 
        & 128 & -- & 348.77 & 8.21 & 64.25 \\
          & 128 & \checkmark & \textbf{317.63} & \textbf{7.77} & \textbf{65.37} \\
\cmidrule{2-6}
          & 256 & -- & 357.95 & 8.28 & 64.55 \\
          & 256 & \checkmark & \textbf{329.39} & \textbf{7.81} & \textbf{65.13} \\
\midrule
\multirow{4}{*}[-2pt]{Qwen3-8B}  & 128 & -- & 704.23 & 12.73 & 63.88 \\
          & 128 & \checkmark & \textbf{638.98} & \textbf{12.29} & \textbf{65.49} \\
\cmidrule{2-6}
          & 256 & -- & 723.86 & 12.58 & 63.70 \\
          & 256 & \checkmark & \textbf{661.31} & \textbf{12.53} & \textbf{64.94} \\
\midrule
\multirow{4}{*}[-2pt]{Qwen3-4B}  & 128 & -- & 370.41 & \textbf{17.40} & 60.87 \\
          & 128 & \checkmark & \textbf{341.48} & 17.53 & \textbf{61.61} \\
\cmidrule{2-6}
          & 256 & -- & 380.76 & 17.69 & 60.80 \\
          & 256 & \checkmark & \textbf{354.16} & \textbf{17.50} & \textbf{61.24} \\
\bottomrule
\end{tabular}
\end{small}
\end{center}
\vspace{-8pt}
\end{table}
\begin{table}[ht]
    \caption{The performance of \methodShort on Llama3-8B under reduced bit-widths. 
    \textbf{N.A.} indicates model divergence or numerical collapse. \textbf{\#Float} denotes the number of floating point digits used as quantization parameters.}
    \vspace{-5pt}
    \label{tab:low_bit_ptq}
    \begin{center}
    \begin{small}
    \begin{tabular}{c l c c c c}
        \toprule
        \textbf{Bit} & \textbf{Method} & \textbf{\#Float} &\textbf{Wiki} $\downarrow$ & \textbf{PTB} $\downarrow$ & \textbf{Avg} $\uparrow$ \\
        \midrule
        \multirow{3}{*}{3} & NormalFloat & 52M & N.A. & N.A. & 33.79 \\
          & LoftQ & 92M & 73.73 & 88.47 & 52.36 \\
          & \methodShort & 50M & \textbf{13.69} & \textbf{27.14} & \textbf{60.76} \\
        \midrule
        \multirow{3}{*}{2.5} & NormalFloat & 52M & N.A. & N.A. & 33.64 \\
            & LoftQ & 92M & 667.30 & 2764.62 & 38.06 \\
            & \methodShort & 50M & \textbf{25.01} & \textbf{50.18} & \textbf{51.03} \\
        \midrule
        \multirow{3}{*}{2.25} & NormalFloat & 52M & N.A. & N.A. & 33.45 \\
             & LoftQ & 92M & N.A. & N.A. & 33.88 \\
             & \methodShort & 50M & \textbf{80.46} & \textbf{170.12} & \textbf{46.54} \\
        \bottomrule
    \end{tabular}
    \end{small}
    \end{center}
    \vspace{-10pt}
\end{table}
\begin{table*}[ht]
    \caption{Comparison of Quantization-Aware Training (QAT) performance between standard block-wise INT4 and \methodShort on Llama3-8B and Qwen3-8B. \textbf{Bold} and \underline{underlined} values indicate the best and second-best performance within each block size group.}
    \vspace{-5pt}
    \label{tab:qat_results}
    \begin{center}
    \begin{small}
    \setlength{\tabcolsep}{3.5pt}
    \begin{tabular}{lccccccccccc}
        \toprule
        \textbf{Model} & \textbf{Method} & \textbf{Wiki} $\downarrow$ & \textbf{PTB} $\downarrow$ & \textbf{BoolQ} & \textbf{PIQA} & \textbf{HS} & \textbf{WG} & \textbf{ARC-e} & \textbf{ARC-c} & \textbf{OBQA} & \textbf{Avg} $\uparrow$ \\
        \midrule
        \rowcolor{gray!10} Llama3-8B & - & 7.26 & 15.84 & 81.28 & 79.49 & 60.23 & 73.09 & 80.13 & 50.43 & 34.80 & 65.64 \\
        \midrule
        \multirow{4}{1.5cm}{\centering block size 128}
         & INT4 & 8.81 & 19.26 & 77.43 & 77.91 & 58.59 & 73.09 & 78.79 & 47.70 & 35.40 & 64.13 \\
         & \methodShort & 8.32 & \underline{18.07} & \textbf{81.22} & \underline{79.16} & 58.72 & 73.32 & \underline{79.55} & \textbf{48.21} & 34.40 & 64.94 \\
         & INT4-QAT & \underline{8.25} & 18.22 & 80.40 & 79.05 & \underline{59.28} & \textbf{74.35} & 79.46 & \underline{48.12} & \underline{34.80} & \underline{65.07} \\
         & \methodShort-QAT & \textbf{8.09} & \textbf{17.72} & \underline{80.92} & \textbf{79.54} & \textbf{59.48} & \underline{73.32} & \textbf{80.26} & 48.04 & \textbf{36.40} & \textbf{65.42} \\
        \cmidrule{2-12}
        \multirow{4}{1.5cm}{\centering block size 256}
         & INT4 & 9.10 & 19.59 & 76.97 & 77.64 & 58.49 & 72.45 & 77.19 & 46.59 & 35.00 & 63.48 \\
         & \methodShort & 8.38 & \underline{18.22} & \textbf{80.28} & 79.00 & 58.68 & 73.01 & \textbf{80.26} & 48.38 & 34.40 & 64.86 \\
         & INT4-QAT & \underline{8.36} & 18.39 & 79.30 & \underline{79.16} & \underline{58.90} & \underline{73.72} & 79.97 & \textbf{49.57} & \underline{35.40} & \underline{65.15} \\
         & \methodShort-QAT & \textbf{8.15} & \textbf{18.11} & \underline{79.82} & \textbf{79.33} & \textbf{59.44} & \textbf{74.35} & \underline{80.05} & \underline{48.63} & \textbf{35.80} & \textbf{65.35} \\
        \midrule
        \rowcolor{gray!10} Qwen3-8B & - & 12.22 & 25.06 & 86.61 & 76.33 & 57.14 & 68.43 & 83.46 & 55.55 & 31.20 & 65.53 \\
        \midrule
        \multirow{4}{1.5cm}{\centering block size 128}
         & INT4 & 14.07 & 31.88 & 85.78 & 75.63 & 55.78 & 65.59 & 79.92 & 52.65 & 29.60 & 63.56 \\
         & \methodShort & 13.42 & 30.83 & 85.08 & 75.84 & \textbf{56.11} & 67.48 & 81.10 & 53.50 & 30.20 & 64.19 \\
         & INT4-QAT & \underline{11.15} & \underline{24.10} & \textbf{85.69} & \underline{77.80} & 55.42 & \underline{71.35} & \textbf{82.87} & \underline{54.44} & \textbf{33.40} & \underline{65.85} \\
         & \methodShort-QAT & \textbf{10.65} & \textbf{22.46} & 84.37 & \textbf{77.97} & \underline{56.09} & \textbf{73.48} & \underline{82.32} & \textbf{55.72} & \underline{32.00} & \textbf{65.99} \\
        \cmidrule{2-12}
        \multirow{4}{1.5cm}{\centering block size 256}
         & INT4 & 14.38 & 33.13 & \textbf{85.93} & 75.30 & 55.68 & 65.75 & 80.18 & 52.99 & 31.40 & 63.89 \\
         & \methodShort & 13.68 & 31.10 & 85.26 & 75.52 & \textbf{56.16} & 66.77 & 80.35 & 52.13 & 30.00 & 63.74 \\
         & INT4-QAT & \underline{11.37} & \underline{24.66} & \underline{85.54} & \underline{77.69} & 55.27 & \underline{71.03} & \underline{81.48} & \underline{54.95} & \underline{32.20} & \underline{65.45} \\
         & \methodShort-QAT & \textbf{10.72} & \textbf{22.62} & 84.83 & \textbf{77.75} & \underline{56.06} & \textbf{72.38} & \textbf{82.45} & \textbf{55.12} & \textbf{32.60} & \textbf{65.88} \\
        \bottomrule
    \end{tabular}
    \end{small}
    \end{center}
    \vspace{-5pt}
\end{table*}
\begin{table*}[ht]
    \caption{Comparison of quantized-PEFT performance on commonsense reasoning benchmarks. Following the experimental protocols of previous works \citep{liu2024dora,hu2023llm}, all models are fine-tuned on the Commonsense-170k dataset for 3 epochs. \textbf{\#Train} denotes the number of trainable parameters, while \textbf{\#Float} represents the total floating-point parameters. }
    \label{tab:peft_results}
    \begin{center}
    \begin{small}
    \setlength{\tabcolsep}{3.5pt}
    \begin{tabular}{lcccccccccccc}
        \toprule
        \textbf{Model} & \textbf{Method} & \textbf{\#Train} & \textbf{\#Float} & \textbf{BoolQ} & \textbf{PIQA} & \textbf{SIQA} & \textbf{HS} & \textbf{WG} & \textbf{ARC-e} & \textbf{ARC-c} & \textbf{OBQA} & \textbf{Avg} \\
        \midrule
        \multirow{3}{*}{\centering Llama3-8B}
         & QLoRA & 84M & 193M & 70.15 & 81.23 & 77.64 & 87.65 & 80.35 & 82.15 & 68.09 & 77.40 & 78.08  \\
         & LoftQ & 84M & 193M & 72.72 & 85.91 & 79.68 & 93.54 & 85.32 & 89.31 & 75.85 & 85.60 & 83.49 \\
         & \methodShort & 84M & \textbf{84M} & 76.39 & 89.72 & 82.55 & 96.38 & 89.58 & 93.39 & 84.47 & 89.00 & \textbf{87.68}  \\
        \midrule
        \multirow{3}{*}{\centering Qwen3-8B}
         & QLoRA & 87M & 196M & 73.98 & 86.89 & 78.61 & 93.46 & 86.98 & 94.61 & 86.01 & 90.40 & 86.37 \\
         & LoftQ & 87M & 196M & 75.38 & 89.55 & 81.27 & 95.24 & 88.00 & 96.17 & 88.74 & 91.6 & 88.24 \\
         & \methodShort & 87M & \textbf{87M} & 75.26 & 90.37 & 81.47 & 95.58 & 89.42 & 97.05 & 91.21 & 91.60 & \textbf{89.00} \\
        \midrule
        \multirow{3}{*}{\centering Qwen3-4B}
         & QLoRA & 66M & 123M & 71.53 & 84.00 & 78.40 & 90.83 & 82.48 & 91.84 & 83.02 & 86.60 & 83.59 \\
         & LoftQ & 66M & 123M & 72.23 & 86.40 & 79.48 & 92.77 & 85.08 & 94.40 & 85.24 & 87.60 & 85.40 \\
         & \methodShort & 66M & \textbf{66M} & 72.75 & 87.05 & 79.68 & 94.01 & 86.58 & 94.82 & 87.20 & 92.00 & \textbf{86.76} \\
        \bottomrule
    \end{tabular}
    \end{small}
    \end{center}
    \vspace{-5pt}
\end{table*}

\textbf{Pushing the Limits: Ultra-Low Bit Quantization } Following LoftQ's setting, we evaluate the robustness of \methodShort at ultra-low bit-widths on Llama3-B ($B=128$).
3/2.5/2.25-bit configurations denote mixed-precision quantization, using NF4 for the first 50\%/25\%/12.5\% of layers and NF2 for the remainder.
For LoftQ, the low-rank adapters are set to a rank of 16. 
The results in Table~\ref{tab:low_bit_ptq} demonstrate that, due to the enhanced expressive power provided by its continuous scaling formulation, the advantage of \methodShort becomes particularly pronounced at ultra-low bit-widths, where it maintains remarkably stable performance while baselines suffer divergence. Compared to the second-best method, our approach achieves accuracy improvements of 8.4\%, 12.97\%, and 12.66\% at 3, 2.5, and 2.25 bits, respectively, despite a lower floating-point budget.

For more results, please refer to  Appendix~\ref{app:quant_error}.

\subsection{\methodShort for Quantization-Aware Training}
\label{exp:qat}

To evaluate the efficacy of \methodShort under Quantization-Aware Training (QAT) setting, we conduct experiments following the experimental protocol established by the \texttt{PyTorch} team~\citep{pytorch2020qat}. 
Our experiments are performed on Llama3-8B and Qwen3-8B using a 4-bit weight-only quantization scheme with equivalent block sizes of 128 and 256. The models are fine-tuned for 1,250 steps on a subset of the SmolLM~\citep{smollm} pretraining corpus with a global batch size of 64. We employ a cosine learning rate scheduler with a linear warmup ratio of 0.3 and a peak learning rate of 2e-5. 
Table \ref{tab:qat_results} confirms that \methodShort consistently benefits from QAT, narrowing the quantization-precision gap. For Qwen3-8B (bs=128), \methodShort-QAT improves the average score from 64.19\% to 65.99\%. Notably, \methodShort-QAT consistently outperforms the INT4-QAT baseline, validating the structural advantage of a continuous scaling manifold over rigid, piecewise-constant block scales.

\begin{figure*}[t]
    \centering
    \centering
    \includegraphics[width=0.9\textwidth]{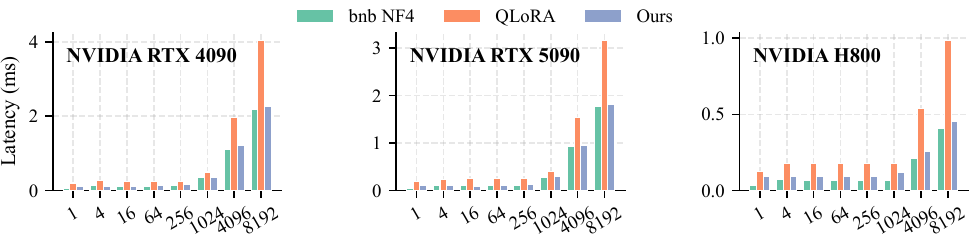}
    \caption{Comparison of operator latency between \texttt{bitsandbytes} NF4, \texttt{peft} QLoRA, and \methodShort (Ours) across various hardware platforms (RTX 4090, RTX 5090, and H800). The x-axis represents the total number of processed tokens $M$. 
    }
    \vspace{-10pt}
    \label{fig:kernel_latency}
\end{figure*}

\subsection{\methodShort for Quantized Parameter-Efficient Fine-Tuning}
\label{exp:peft}

In this section, we evaluate \methodShort as a quantized Parameter-Efficient Fine-Tuning (PEFT) framework against various baselines, including QLoRA \citep{dettmers2023qlora} and LoftQ \citep{li2023loftq}.
Following the experimental protocols established in previous work~\citep{liu2024dora,hu2023llm}, we conduct fine-tuning on the Commonsense-170k dataset and evaluate performance across eight distinct commonsense reasoning benchmarks: BoolQ, PIQA, SIQA, HellaSwag, WinoGrande, ARC-e, ARC-c, and OpenBookQA.

Following the training recipe from DoRA, all models are fine-tuned for 3 epochs with a global batch size of 16 and a linear learning rate scheduler. For \methodShort, we adopt a peak learning rate of $1 \times 10^{-4}$. 
All experiments are conducted using 4-bit NF4 quantization with a block size of 64 and adapters' rank as 32. Since the quantization and adaptation processes in \methodShort occur synchronously, we apply our low-rank scaling adapters to all linear layers within each transformer block.

The experimental results summarized in Table~\ref{tab:peft_results} demonstrate the superior performance of \methodShort across multiple model architectures and benchmarks. 
Specifically, \methodShort consistently outperforms prominent baselines such as QLoRA and LoftQ in all evaluated settings. 
For instance, on the Llama3-8B model, \methodShort achieves an average accuracy of \textbf{87.68\%}, surpassing LoftQ by a significant margin of \textbf{4.19\%} while using substantially fewer floating-point parameters (84M vs. 193M). 
A similar trend is observed in the Qwen3 series, where \methodShort establishes new state-of-the-art performance levels for quantized adaptation.

To empirically verify the expressivity of \methodShort, we also perform rank analysis on the weight update matrix $\Delta W$. As detailed in Appendix~\ref{app:rank_visualization}, \methodShort achieves full-rank updates with a smooth, long-tail singular value distribution.

\subsection{Computational Efficiency and Kernel Performance}
\label{exp:kernel}

Unlike established quantized-PEFT methods such as QLoRA, which introduce auxiliary full-precision additive adapters that cannot be merged, \methodShort maintains a consistent model architecture throughout its entire lifecycle. 
To take full advantage of this theoretical advantage, we implement highly optimized \texttt{Triton} kernels for \methodShort inference.

We first conduct a micro-benchmark focusing on the first \texttt{q\_proj} layer of Llama3-8B. We compare the latency of our kernel against the standard NF4 implementation in \texttt{bitsandbytes} and the additive QLoRA adapter. As shown in Figure~\ref{fig:kernel_latency}, across various GPUs (RTX 4090, RTX 5090, and NVIDIA H800), \methodShort consistently demonstrates superior efficiency. On the H800 with a sequence length of 8192, our method (0.456ms) improves throughput to 2.17x comparing to QLoRA (0.987ms) and even matches the industrial-level NF4 implementation (0.409ms).

We also measure end-to-end throughput on Llama3-8B with a batch size of 2, an input length of 4096, and an output length of 128 tokens. The results, summarized in Table~\ref{tab:throughput}, indicate that \methodShort enjoys a consistently higher throughput than QLoRA on all hardware platforms.

\begin{table}[ht]
\centering
\caption{End-to-end inference throughput comparison on Llama3-8B. Measurements were taken with a batch size of 2, input length of 4096, and output length of 128 tokens.}
\label{tab:throughput}
\begin{small}
\begin{tabular}{llccc}
\toprule
\textbf{Machine} & \textbf{Method} & \textbf{Prefill} & \textbf{Decode} & \textbf{Total} \\
 & & (tokens/s) & (tokens/s) & (tokens/s) \\
\midrule
\multirow{3}{*}{RTX 4090} & \texttt{bnb} NF4 & 8163.39 & 47.48 & 1320.90 \\
 & QLoRA & 5601.44 & 27.58 & 786.35 \\
 & \methodShort & 7847.91 & 43.39 & 1216.69 \\
\midrule
\multirow{3}{*}{RTX 5090} & \texttt{bnb} NF4 & 11327.05 & 61.75 & 1735.17 \\
 & QLoRA & 8267.42 & 35.99 & 1042.47 \\
 & \methodShort & 10951.46 & 45.08 & 1314.46 \\
\midrule
\multirow{3}{*}{H800} & \texttt{bnb} NF4 & 30278.60 & 63.91 & 1975.68 \\
 & QLoRA & 17703.85 & 35.11 & 1089.49 \\
 & \methodShort & 27700.28 & 47.65 & 1490.49 \\
\bottomrule
\end{tabular}
\end{small}
\vspace{-15pt}
\end{table}

\section{Conclusion}
\label{sec:conclusion}

In this paper, we introduced \methodFull, a versatile framework that reconceptualizes the scaling mechanism in Large Language Model quantization. By interpolating between block-wise and element-wise quantization with a continuous low-rank manifold via SVD, we provide a unified optimization space for the entire model lifecycle. 
We also bridge the gap between theoretical innovation and practical deployment by developing highly optimized \texttt{Triton} kernels.
Our experiments demonstrate that \methodShort not only achieves superior reconstruction accuracy in PTQ and QAT settings across various model families but also enables a highly expressive, zero-overhead PEFT paradigm. By moving beyond the limitations of piecewise-constant scaling, \methodShort establishes a new standard for balancing compression efficiency with model performance.

The title of this work, ``\textit{Breaking the Blocks},'' carries a dual significance that encapsulates our core philosophy. On a technical level, we break the spatial constraints of block-wise quantization, replacing static, discrete grids with a fluid and continuous scaling representation. On a conceptual level, we break the blocks that traditionally separate the stages of model compression and adaptation. By unifying PTQ, QAT, and PEFT under a single low-rank formulation, we provide a seamless path from the initial quantization of a pre-trained model to its final deployment-ready, task-specific state. We hope this work encourages a more integrated view of model efficiency, where compression and adaptation are treated not as disparate hurdles, but as a single, synergistic optimization process.

\newpage
\section*{Impact Statement}

This paper presents work whose goal is to advance the field
of Machine Learning. There are many potential societal
consequences of our work, none which we feel must be
specifically highlighted here.

\bibliography{custom}
\bibliographystyle{icml2026}

\newpage
\appendix
\onecolumn

\section{Detailed Rank Configurations and Parameter Alignment}
\label{app:rank}

To align the number of parameters in the A and B matrices with that of the scale matrix, the rank for each module is determined dynamically based on the matrix shape ($m \times n$) and the block size $B$, using the formula $r = \lfloor \frac{nm}{B(n + m)} \rfloor$. The resulting ranks for all modules are listed in Table \ref{tab:rank_results}.

\begin{table}[ht]
\centering
\caption{The ranks of the $A$ and $B$ matrices under various block size configurations for each model and module.}
\label{tab:rank_results}
\begin{tabular}{lccccc}
\toprule
\multirow{3}{*}{\textbf{Model}} & \multirow{3}{*}{\textbf{Module}} & \multirow{3}{*}{\textbf{shape}} & \multicolumn{2}{c}{\textbf{Blocksize}} \\
\cmidrule{4-5}
 & & & \textbf{128} & \textbf{256} \\
\midrule
\multirow{4}{*}{Llama3-8B} 
 & Q/O & $4096 \times 4096$ & 16 & 8  \\
 & K/V & $1024 \times 4096$ & 6 & 3    \\
 & Up/Gate & $14336 \times 4096$ & 24 & 12    \\
 & Down & $4096 \times 14336$ & 24 & 12    \\
\midrule
\multirow{4}{*}{Qwen3-8B} 
 & Q/O & $4096 \times 4096$ & 16 & 8  \\
 & K/V & $1024 \times 4096$ & 6 & 3    \\
 & Up/Gate & $12288 \times 4096$ & 24 & 12    \\
 & Down & $4096 \times 12288$ & 24 & 12    \\
\midrule
\multirow{5}{*}{Qwen3-4B} 
 & Q & $4096 \times 2560$ & 12 & 6  \\
 & O & $2560 \times 4096$ & 12 & 6  \\
 & K/V & $1024 \times 2560$ & 5 & 2    \\
 & Up/Gate & $9728 \times 2560$ & 15 & 7    \\
 & Down & $2560 \times 9728$ & 15 & 7    \\
\bottomrule
\end{tabular}
\end{table}
\section{Comparing the Quantization Error of NF4, LoftQ , QPiSSA and \methodShort in different models and lower bits}
\label{app:quant_error}
This section compares \methodShort against standard NF4, LoftQ, and QPiSSA in terms of quantization error. The experiments are conducted using the setup detailed in Section~\ref{exp:ptq}. For \methodShort, the rank of each module is assigned dynamically, and the model is trained for 500 steps with a learning rate of 0.05. For LoftQ and QPiSSA, we use an adapter of rank 16 and perform 5 iterations. To ensure a fair comparison of parameter efficiency with LoRA-based methods, we also train a parameter-aligned variant of \methodShort. In this configuration, the rank calculation for \methodShort accounts for the size of both the scale matrix and the adapter matrix, as given by the formula $r = \lfloor \frac{nm}{B(n + m)} \rfloor + r_q$, where $r_q$ denotes the rank of the adapter used in LoRA-based methods. We adopt the \textbf{Quantization Error Reduction Ratio} as the evaluation metric, defined as $1-\frac{\| W - Q \|_*}{\| W - \text{nf4}(W) \|_*}$. A \textit{higher} value for this rate corresponds to \textit{lower} quantization error and better model performance. The results are summarized in Table~\ref{tab:quant_error_ratio_results}. It can be observed that \methodShort already achieves a lower quantization error than LoftQ and QPiSSA even without parameter alignment. Furthermore, under the parameter-aligned setting, \methodShort establishes a substantially larger performance advantage over the other methods.

Table~\ref{tab:quant_error_ratio_lowbit} presents a comparison of the quantization error between \methodShort and other methods under lower bit-width settings. Here, 2-bit indicates the use of NF2 quantization, while the 3-bit, 2.5-bit, and 2.25-bit configurations correspond to mixed-precision schemes where the initial 50\%, 25\%, and 12.5\% of the model's layers are quantized using NF4, respectively, and the remaining layers are quantized with NF2. All experiments are conducted on the Llama3-8B model using a block size of 128. The training steps and learning rates follow the low-bit quantization settings described in Section~\ref{exp:ptq}. The quantization error reduction ratio is adopted as the evaluation metric. The results demonstrate that \methodShort achieves a quantization error reduction ratio approximately three times higher than other methods in low-bit scenarios. Moreover, as the bit-width further decreases, \methodShort exhibits a progressively increasing quantization error reduction rate—rising from 31.8\% at 3-bit to 35.7\% at 2-bit.

\begin{table}[ht]
\centering
\caption{The quantization error reduction ratio of \methodShort and other compared methods across various model modules. The annotation $\text{\methodShort}^\dag$ denotes the parameter-aligned version of \methodShort, which is designed for a fair comparison with LoRA-based methods.}
\small
\label{tab:quant_error_ratio_results}
\begin{tabular}{cccccccccccccc}
\toprule
\textbf{Model} & \textbf{Method}  & \textbf{Blocksize} & \textbf{\#Float.} & Q & K & V & O & Gate & Up & Down & \textbf{AVG}$\uparrow$  \\
\midrule
\multirow{10}{*}{Llama3-8B } 
& NF4 & 128 & 52M & 0 & 0 & 0 & 0 & 0 & 0 & 0 & 0  \\
& LoftQ & 128 & 92M & 3.4 & 6.8 & 6.7 & 3.4 & 2.3 & 2.4 & 2.3 & 3.0 \\
& QPiSSA & 128 & 92M & 7.2 & 11.3 & 6.5 & 4.0 & 3.4 & 2.5 & 2.5 & 4.3 \\
& \methodShort & - & 50M & 7.7 & 6.8 & 6.4 & 6.1 & 7.7 & 6.4 & 7.5 & 7.1 \\ 
& $\text{\methodShort}^\dag$ & - & 90M & \textbf{12.2} & \textbf{14.7} & \textbf{15.3} & \textbf{9.8} & \textbf{10.1} & \textbf{8.8} & \textbf{9.7} & \textbf{10.5} \\
\cmidrule{2-12}
& NF4 & 256 & 26M  & 0 & 0 & 0 & 0 & 0 & 0 & 0 & 0  \\
& LoftQ & 256 & 66M & 3.3 & 6.7 & 6.6 & 3.4 & 2.3 & 2.3 & 2.2 & 2.9 \\
& QPiSSA & 256 & 66M & 7.5 & 11.6 & 6.5 & 4.1 & 3.5 & 2.5 & 2.5 & 4.4 \\
& \methodShort & - & 25M & 7.9 & 8.9 & 6.9 & 5.3 & 6.7 & 5.6 & 6.1 & 6.5 \\
& $\text{\methodShort}^\dag$ & - & 65M & \textbf{14.3} & \textbf{18.9} & \textbf{18.0} & \textbf{12.0} & \textbf{11.0} & \textbf{9.9} & \textbf{10.6} & \textbf{12.0} \\
\midrule
\multirow{10}{*}{Qwen3-8B } 
& NF4 & 128 & 51M  & 0 & 0 & 0 & 0 & 0 & 0 & 0 & 0  \\
& LoftQ & 128 & 93M & 3.2 & 6.4 & 6.8 & 3.4 & 2.4 & 2.5 & 2.7 & 3.0 \\
& QPiSSA & 128 & 93M & 5.7 & 9.3 & 6.5 & 3.8 & 3.7 & 2.5 & 3.1 & 3.8 \\
& \methodShort & - & 51M & 6.6 & 6.1 & 6.9 & 5.0 & 7.0 & 6.6 & 7.9 & 6.8 \\ 
& $\text{\methodShort}^\dag$ & - & 93M & \textbf{10.1} & \textbf{13.4} & \textbf{14.0} & \textbf{9.2} & \textbf{10.1} & \textbf{9.8} & \textbf{11.0} & \textbf{10.4} \\
\cmidrule{2-12}
& NF4 & 256 & 26M  & 0 & 0 & 0 & 0 & 0 & 0 & 0 & 0  \\
& LoftQ & 256 & 68M & 3.1 & 6.4 & 6.6 & 3.3 & 2.4 & 2.4 & 2.6 & 2.9 \\
& QPiSSA & 256 & 68M & 5.9 & 9.6 & 6.4 & 3.8 & 3.9 & 2.6 & 3.1 & 3.9 \\
& \methodShort & - & 26M & 8.0 & 7.8 & 8.0 & 6.2 & 7.7 & 6.9 & 8.6 & 7.6 \\
& $\text{\methodShort}^\dag$ & - & 67M & \textbf{12.5} & \textbf{16.7} & \textbf{17.1} & \textbf{11.4} & \textbf{11.7} & \textbf{11.1} & \textbf{12.7} & \textbf{12.2} \\
\midrule
\multirow{10}{*}{Qwen3-4B } 
& NF4 & 128 & 27M  & 0 & 0 & 0 & 0 & 0 & 0 & 0 & 0  \\
& LoftQ & 128 & 59M & 4.1 & 7.3 & 7.6 & 4.2 & 3.4 & 3.5 & 3.4 & 4.0 \\
& QPiSSA & 128 & 59M & 6.7 & 10.2 & 7.4 & 4.6 & 4.5 & 3.3 & 3.7 & 4.7 \\
& \methodShort & - & 26M & 5.9 & 4.3 & 5.2 & 4.5 & 5.2 & 5.5 & 6.9 & 5.6 \\ 
& $\text{\methodShort}^\dag$ & - & 57M & \textbf{10.0} & \textbf{13.3} & \textbf{13.6} & \textbf{9.6} & \textbf{9.9} & \textbf{10.3} & \textbf{11.7} & \textbf{10.6} \\
\cmidrule{2-12}
& NF4 & 256 & 14M  & 0 & 0 & 0 & 0 & 0 & 0 & 0 & 0  \\
& LoftQ & 256 & 45M & 4.1 & 7.3 & 7.5 & 4.1 & 3.3 & 3.5 & 3.3 & 3.9 \\
& QPiSSA & 256 & 45M & 6.7 & 10.3 & 7.4 & 4.7 & 4.6 & 3.4 & 3.7 & 4.8 \\
& \methodShort & - & 12M & 6.7 & 4.2 & 4.4 & 6.0 & 5.3 & 5.2 & 7.4 & 5.9 \\
& $\text{\methodShort}^\dag$ & - & 44M & \textbf{12.3} & \textbf{15.8} & \textbf{16.1} & \textbf{12.2} & \textbf{11.1} & \textbf{11.4} & \textbf{13.4} & \textbf{12.4} \\
\bottomrule
\end{tabular}
\end{table}

\begin{table}[ht]
\centering
\caption{The quantization error reduction ratio of \methodShort in low bit setting.}
\label{tab:quant_error_ratio_lowbit}
\small
\begin{tabular}{cccccccccccccc}
\toprule
\textbf{Bit} & \textbf{Method}& \textbf{\#Float.}  & Q & K & V & O & Gate & Up & Down & \textbf{AVG}$\uparrow$  \\
\midrule
\multirow{4}{*}{3 } 
& NF4 & 52M & 0 & 0 & 0 & 0 & 0 & 0 & 0 & 0  \\
& LoftQ & 92M & 10.0 & 12.1 & 11.6 & 9.1 & 9.3 & 8.9 & 8.7 & 9.4 \\
& QPiSSA & 92M & 12.1 & 14.6 & 11.6 & 9.7 & 9.8 & 9.0 & 8.9 & 10.1\\
& \methodShort & 50M & \textbf{32.4} & \textbf{32.7} & \textbf{33.1} & \textbf{31.9} & \textbf{31.9} & \textbf{31.3} & \textbf{31.3} & \textbf{31.8} \\ 
\midrule
\multirow{4}{*}{2.5 } 
& NF4 & 52M & 0 & 0 & 0 & 0 & 0 & 0 & 0 & 0  \\
& LoftQ & 92M & 10.8 & 12.6 & 11.8 & 9.7 & 10.0 & 9.5 & 9.4 & 10.1 \\
& QPiSSA & 92M & 12.7 & 14.9 & 11.9 & 10.4 & 10.6 & 9.7 & 9.7 & 10.8 \\
& \methodShort & 50M & \textbf{35.0} & \textbf{35.3} & \textbf{34.9} & \textbf{34.2} & \textbf{34.3} & \textbf{33.9} & \textbf{33.9} & \textbf{34.3} \\ 
\midrule
\multirow{4}{*}{2.25 } 
& NF4 & 52M & 0 & 0 & 0 & 0 & 0 & 0 & 0 & 0  \\
& LoftQ & 92M & 11.0 & 12.8 & 11.8 & 9.8 & 10.2 & 9.7 & 9.6 & 10.3 \\
& QPiSSA & 92M & 12.9 & 15.0 & 12.0 & 10.6 & 10.8 & 9.9 & 10.0 & 11.0 \\
& \methodShort & 50M & \textbf{35.9} & \textbf{36.1} & \textbf{35.5} & \textbf{35.0} & \textbf{35.0} & \textbf{34.7} & \textbf{34.7} & \textbf{35.1} \\ 
\midrule
\multirow{4}{*}{2 } 
& NF4 & 52M & 0 & 0 & 0 & 0 & 0 & 0 & 0 & 0  \\
& LoftQ & 92M & 11.3 & 13.1 & 11.9 & 10.0 & 10.3 & 9.9 & 9.8 & 10.5 \\
& QPiSSA & 92M & 13.3 & 15.3 & 12.0 & 10.8 & 10.9 & 10.0 & 10.2 & 11.2 \\
& \methodShort & 50M & \textbf{36.6} & \textbf{36.8} & \textbf{35.9} & \textbf{35.6} & \textbf{35.6} & \textbf{35.3} & \textbf{35.4} & \textbf{35.7} \\ 
\bottomrule
\end{tabular}
\end{table}
\section{Visualization of Rank Updates}
\label{app:rank_visualization}

To empirically validate our claim that \methodShort enables high-rank weight updates despite its low-rank parameter budget, we analyze the singular value distribution of the weight update matrix $\Delta W$.

\begin{figure}[h]
    \centering
    \begin{subfigure}[b]{0.48\linewidth}
        \centering
        \includegraphics[width=0.7\linewidth]{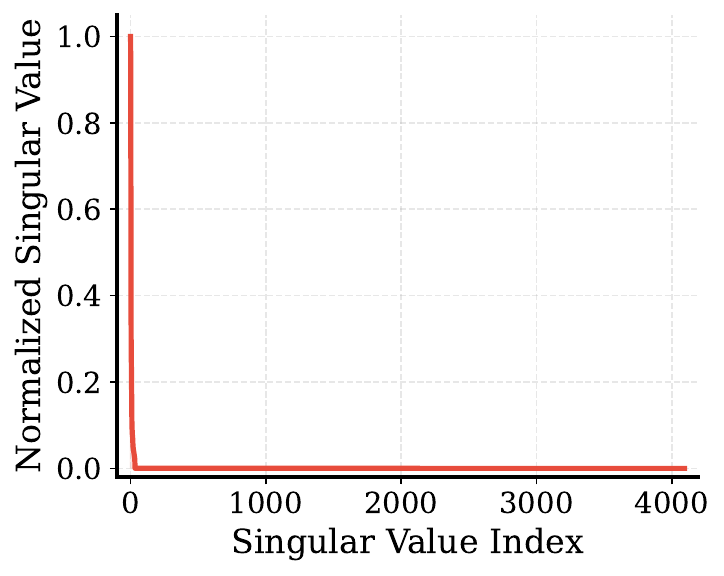}
        \caption{QLoRA}
        \label{fig:svd_qlora}
    \end{subfigure}
    \hfill
    \begin{subfigure}[b]{0.48\linewidth}
        \centering
        \includegraphics[width=0.7\linewidth]{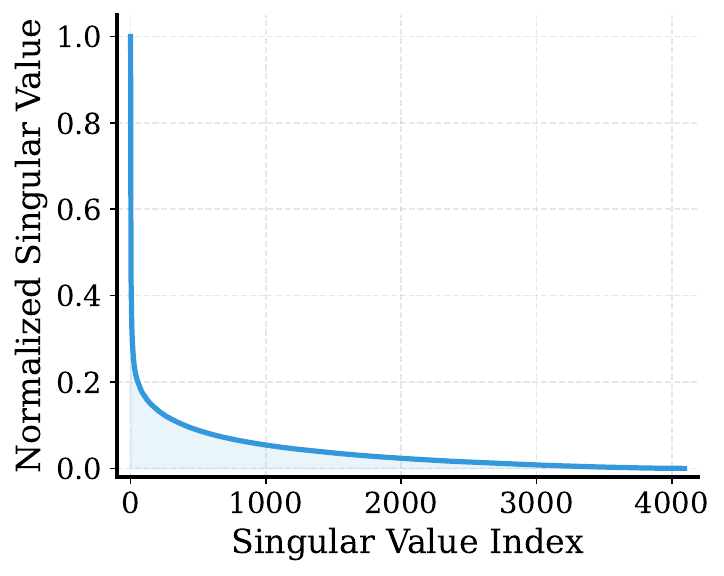}
        \caption{\methodShort (Ours)}
        \label{fig:svd_lords}
    \end{subfigure}
    \caption{Singular value distribution of the weight update $\Delta W$ for the first $Q$ projection layer of Llama3-8B. While the rank of QLoRA is strictly constrained by its additive design, \methodShort achieves full-rank updates similar to full fine-tuning through its multiplicative scaling manifold.}
    \label{fig:svd_comparison}
\end{figure}

Specifically, we extract the weights of the first \texttt{q\_proj} layer from the Llama3-8B model fine-tuned using \methodShort and QLoRA, respectively. We then compute the difference matrix $\Delta W$ and perform Singular Value Decomposition (SVD) to obtain its singular values. As illustrated in Figure \ref{fig:svd_comparison}, the singular values of QLoRA exhibit a sharp truncation, dropping to zero immediately after the predefined rank $r=32$. This behavior is a direct consequence of the additive nature of LoRA-based adapters, which constrains the update to a strictly low-rank subspace.

In contrast, the singular values of \methodShort exhibit a smooth, long-tail distribution that spans the entire dimension of the weight manifold. This phenomenon occurs because \methodShort implements adaptation via a multiplicative scaling mechanism, where the Hadamard product of the full-rank pre-trained matrix $W$ and the low-rank scaling factors effectively ``unfolds'' the update across a high-rank space. This visualization provides strong evidence that \methodShort possesses superior expressive power compared to additive counterparts, enabling it to capture more nuanced adaptation patterns within the same parameter constraints.

\section{Declaration of AI Usage}

Generative AI tools were used for grammar refinement and language polishing to enhance the readability of the manuscript. AI assistance was also employed during the coding and implementation phases of the project. All AI-assisted outputs were reviewed by the authors to ensure the technical quality and accuracy of the final paper.
\end{document}